\documentclass{article} % For LaTeX2e
\usepackage{iclr2026_conference,times}

\usepackage{amsmath,amsfonts,bm}

\def\eqref#1{equation~\ref{#1}}
\def\1{\bm{1}}

\DeclareMathAlphabet{\mathsfit}{\encodingdefault}{\sfdefault}{m}{sl}
\SetMathAlphabet{\mathsfit}{bold}{\encodingdefault}{\sfdefault}{bx}{n}

\newcommand{\Point}{\mathbf{P}}
\newcommand{\Time}{t}
\newcommand{\View}{v}
\newcommand{\nView}{V}
\newcommand{\RGBImage}{I}

\newcommand{\RGBVideo}{\mathbf{I}}
\newcommand{\MaskVideo}{\mathbf{M}}

\newcommand{\Camera}{\mathbf{C}}
\newcommand{\videoLength}{L}

\newcommand{\latent}{\mathbf{z}}
\newcommand{\allLatent}{\mathbf{Z}}
\newcommand{\gaussians}{\mathbf{G}}
\newcommand{\plucker}{\mathbf{E}}
\newcommand{\vaeEncoder}{\mathcal{E}}

\newcommand{\rgbDecoder}{\mathcal{D}_{rgb}}
\newcommand{\gsDecoder}{\mathcal{D}_{s}}
\newcommand{\dynamicgsDecoder}{\mathcal{D}_{s}}
\newcommand{\videoDiffusionModel}{\mathcal{V}}

\newcommand{\btimerbaseline}{BTimer~(GEN3C)}

\usepackage{hyperref}
\usepackage{url}

\usepackage[utf8]{inputenc} % allow utf-8 input
\usepackage[T1]{fontenc}    % use 8-bit T1 fonts
\usepackage{hyperref}       % hyperlinks
\usepackage{url}            % simple URL typesetting
\usepackage{booktabs}       % professional-quality tables
\usepackage{amsfonts}       % blackboard math symbols
\usepackage{nicefrac}       % compact symbols for 1/2, etc.
\usepackage{microtype}      % microtypography
\usepackage{xcolor}         % colors

\usepackage[pdftex]{graphicx}
\usepackage{floatrow}
\usepackage{comment}
\usepackage{bbding}
\usepackage{tabularx}
\usepackage{colortbl}
\usepackage{multirow}
\usepackage{xspace}
\usepackage[acronym]{glossaries}
\usepackage{float}
\usepackage{wrapfig}
\usepackage{algorithm}
\usepackage{algpseudocode}
\usepackage{booktabs}
\usepackage{colortbl}
\usepackage{makecell}
\usepackage{gensymb}
\usepackage{mwe}
\usepackage{subcaption}
\usepackage{enumitem}

\floatstyle{plaintop}
\restylefloat{table}
\usepackage{caption}

\usepackage{soul}
\setuldepth{foobar}

\usepackage{soul} % For colored underline.
\usepackage{bm}
\usepackage{amssymb} % for \triangleq
\definecolor{codegreen}{rgb}{0,0.6,0}
\definecolor{codegray}{rgb}{0.5,0.5,0.5}
\definecolor{codepurple}{rgb}{0.58,0,0.82}
\definecolor{backcolour}{rgb}{0.95,0.95,0.92}
\definecolor{mediumtealblue}{rgb}{0.0, 0.33, 0.71}
\definecolor{darkpastelgreen}{rgb}{0.01, 0.75, 0.24}
\definecolor{azure}{rgb}{0.0, 0.5, 1.0}

\newcommand{\inlinesection}[1]{\noindent{\textbf{#1}}}

\newcommand{\methodnamestr}{Lyra}
\newcommand{\methodname}{\emph{\methodnamestr}}

\author{%
Sherwin Bahmani$^{1,2,3}$ \; Tianchang Shen$^{1,2,3}$ \; Jiawei Ren$^{1}$ \; Jiahui Huang$^{1}$ \; \textbf{Yifeng Jiang}$^{1}$ \\[2pt]
\textbf{Haithem Turki}$^{1}$ \; \textbf{Andrea Tagliasacchi}$^{2,4}$ \; \textbf{David B. Lindell}$^{2,3}$ \; \textbf{Zan Gojcic}$^{1}$ \\[2pt]
\textbf{Sanja Fidler}$^{1,2,3}$ \; \textbf{Huan Ling}$^{1}$ \; \textbf{Jun Gao}$^{1*}$ \; \textbf{Xuanchi Ren}$^{1,2,3*}$ \\[2pt]
$^{1}$NVIDIA\; $^{2}$University of Toronto\; $^{3}$Vector Institute\; $^{4}$Simon Fraser University \\
\small{*equal contribution} \\
}

\title{\methodnamestr: Generative 3D Scene Reconstruction via Video Diffusion Model Self-Distillation}

\iclrfinalcopy % Uncomment for camera-ready version, but NOT for submission.
\begin{document}

\maketitle
\vspace{-3.5em}
\begin{center}
    \url{https://research.nvidia.com/labs/toronto-ai/lyra}
\end{center}
\maketitle
\vspace{0mm}
\begin{center}
\centering
\vspace{-0.3cm}
%\vspace{-0.8cm}
\includegraphics[width=5.5in]{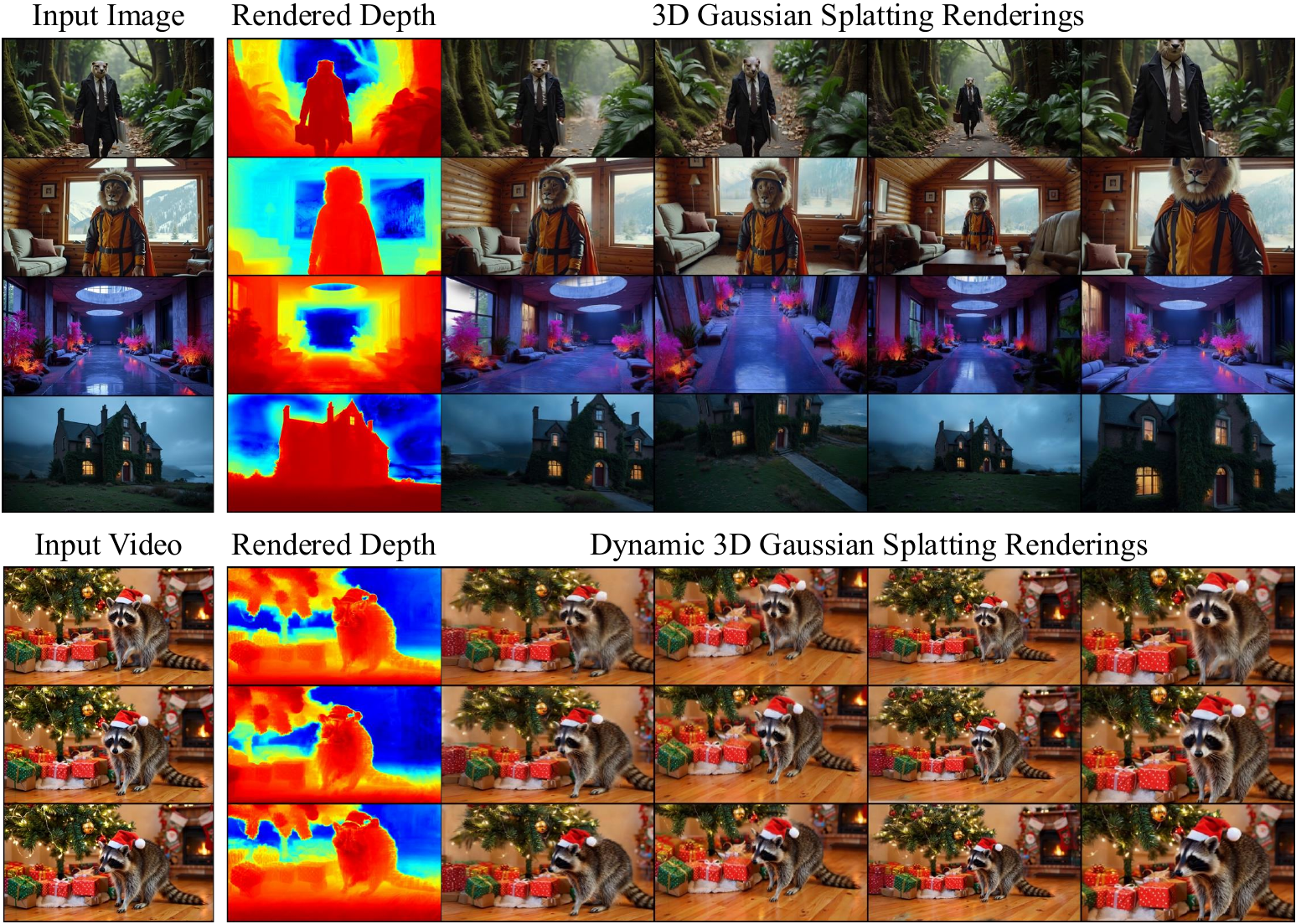}
\captionof{figure}{\textbf{Feed-Forward 3D and 4D Scene Generation.}
From a single image (top), \methodname\ infers a 3D Gaussian Splatting (3DGS) representation in a feed-forward fashion, through self-distilling a video diffusion model without requiring real-world multi-view data.
With a video input (bottom), \methodname\ infers a dynamic 3DGS that offers interactive control in both time (rows) and viewpoint (columns).
}
\label{fig:teaser}
\end{center}

\maketitle
\begin{abstract} 
The ability to generate virtual environments is crucial for applications ranging from gaming to physical AI domains such as robotics, autonomous driving, and industrial AI.
Current learning-based 3D reconstruction methods rely on the availability of captured real-world multi-view data, which is not always readily available. 
Recent advancements in video diffusion models have shown remarkable imagination capabilities, yet their 2D nature limits the applications to simulation where a robot needs to navigate and interact with the environment.
In this paper, we propose a self-distillation framework that aims to distill the implicit 3D knowledge in the video diffusion models into an explicit  3D Gaussian Splatting (3DGS) representation, eliminating the need for multi-view training data.
Specifically, we augment the typical RGB decoder with a 3DGS decoder, which is supervised by the output of the RGB decoder. 
In this approach, the 3DGS decoder can be purely trained with synthetic data generated by video diffusion models.
At inference time, our model can synthesize 3D scenes from either a text prompt or a single image for real-time rendering. 
Our framework further extends to dynamic 3D scene generation from a monocular input video.
Experimental results show that our framework achieves state-of-the-art performance in static and dynamic 3D scene generation.
\end{abstract}
    
\vspace{-3mm}
\section{Introduction}
\label{sec:introduction}
\vspace{-2mm}

Creating high-quality 3D environments at scale is a long-standing challenge in computer vision and graphics,  empowering applications across film-making, VR/AR, and physical AI with closed-loop simulation. These applications require explicit 3D representations that support real-time rendering, physical interaction, and consistent multi-view synthesis.

Recent advances in neural 3D reconstruction~\citep{NeRF, kerbl20233d, 3dgrt2024} enable recovering such representations from posed images.
However, the reliance on accurate camera poses and high-quality images significantly limits their scalability. The challenge is even greater for dynamic scenes, which typically require synchronized multi-camera setups~\citep{li2022neural}. Reconstruction methods are also inherently constrained to the observed content and cannot extrapolate beyond the input views. 
Recent works aim to mitigate these limitations by using feed-forward reconstruction models~\citep{zhang2024gs,ren2024scube}, but these approaches face their own bottleneck: the scarcity of diverse, large-scale 3D training data, which leads to poor out-of-domain generalization. 

Video diffusion models~\citep{agarwal2025cosmos,wan2025wan} offer a promising alternative. Trained on massive internet-scale video corpora, they achieve impressive fidelity and generalization across diverse environments. By learning from real-world videos with varied camera trajectories, ranging from handheld motion to cinematic panning and drone footage, these models implicitly encode cues about the underlying 3D world without requiring multi-view training data.
Unlike reconstruction approaches~\citep{charatan2024pixelsplat}, generative models can also hallucinate plausible content beyond what is visible in the input frames. 
Yet, video diffusion models generate only 2D frames, lacking explicit 3D representations. This limits their use in simulation and downstream tasks that demand geometric consistency, long-term coherence, and physical interaction. 
Encouragingly, recent work~\citep{MotionCtrl, ren2025gen3c}  has shown that video models can be adapted for explicit camera control using relatively small-scale posed datasets. This ability to generate posed image sequences transforms them into a powerful tool that can serve as both input and supervision for 3D reconstruction models.

\begin{figure}[t]
\centering
\vspace{-4mm}
\includegraphics[width=\textwidth]{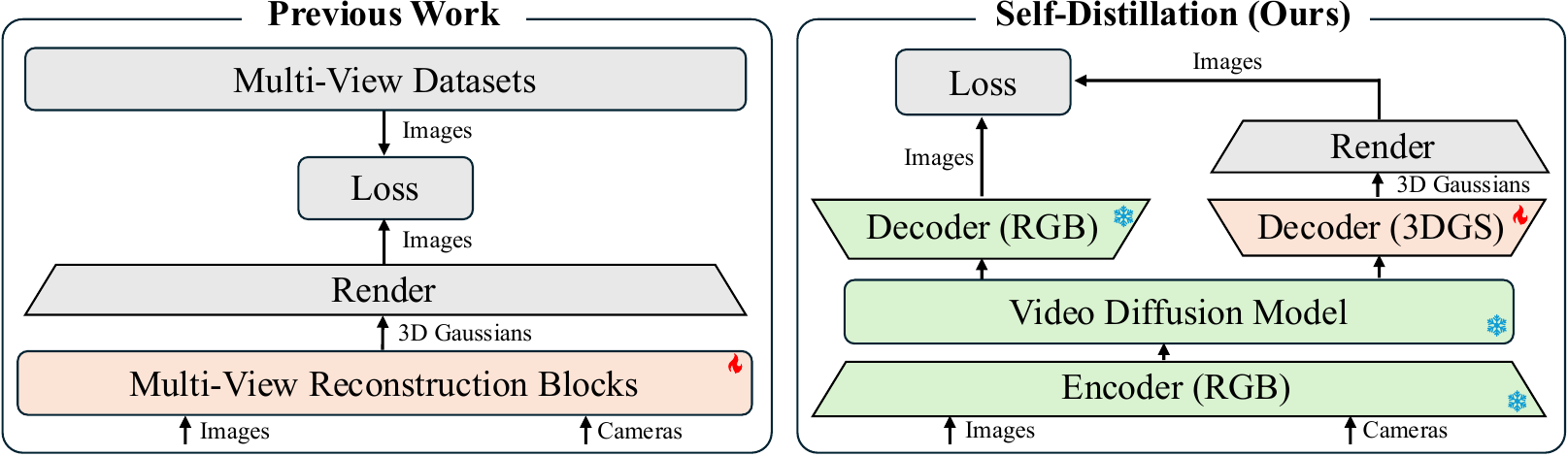}
\caption{\textbf{Self-distillation framework of \methodname.}
Previous work (left)~\citep{szymanowicz2025bolt3d} trains multi-view reconstruction blocks using real-world datasets with limited diversity~\citep{RealEstate10k, ling2024dl3dv}. In contrast, we propose a self-distillation framework (right) for generative 3D scene reconstruction. Precisely, a pre-trained camera-controlled video diffusion model with its RGB decoder output (teacher) supervises the rendering of the 3DGS decoder (student). Using a video model pre-trained on diverse 2D video data allows us to provide diverse multi-view supervision.
}
\label{fig:self_distillation}
\end{figure}

Motivated by this insight, we bridge these two paradigms, reconstruction and generation, through what we call \emph{generative 3D scene reconstruction}. 
We introduce \methodname, a novel method for generating explicit 3D environments from the latent representations of video diffusion models in a single forward pass. 
Cruically, \methodname~is trained in a self-distillation framework (Fig.~\ref{fig:self_distillation}), with a 3D Gaussian Splatting (3DGS) decoder operating directly in the latent space of a video diffusion model.
Given a single input image or video, we sample a camera trajectory as conditioning input, denoise the resulting video latent, and decode it along two parallel branches. The first branch uses the standard RGB decoder to synthesize a video sequence, while the second is our 3DGS decoder, which produces an explicit 3D representation. Together, these branches form a self-distillation framework in which the RGB branch (teacher) supervises the 3DGS branch (student). To extend viewpoint coverage, we sample multiple camera trajectories, generate multiple video latents, and train the 3DGS decoder to fuse information across them while mitigating long-term and multi-view inconsistencies. 

This generative reconstruction framework provides three key benefits: \textbf{(i)} it enables generation of large-scale synthetic environments spanning diverse scenarios directly from video diffusion models, removing the need for real-world multi-view captures; \textbf{(ii)} by operating in latent space of a video model, it allows efficient processing of multiple views without the heavy memory overhead of pixel-space feed-forward reconstruction methods; and \textbf{(iii)} its explicit 3DGS output guarantees geometric consistency, providing representations that are directly applicable to downstream tasks such as physical simulation. 
Consequently, at inference time, Lyra generates high-quality 3D Gaussian scenes from monocular input that support real-time rendering, \textit{without requiring any additional optimization or post-processing}.

% Extension to 4D
We further extend this self-distillation framework to dynamic 4D generation from monocular video. In this setting, the video model (teacher) provides space–time supervision, while the student learns to produce time-conditioned 3DGS representations that enable novel-view synthesis of dynamic scenes.

Overall, our work makes the following contributions:
\begin{itemize}[itemsep=-.1em, ,topsep=0pt,leftmargin=*]
    \item We introduce a self-distillation framework that trains a 3DGS student decoder using a pre-trained camera-controlled video diffusion model as the teacher, eliminating the need for captured multi-view real-world data.
    \item Our framework extends to dynamic scenes to support 4D reconstruction from a monocular video.
    \item Our model generalizes across diverse scenes, achieving state-of-the-art results in single-image 3D scene generation and single-video 4D scene generation.
\end{itemize}

\vspace{-2mm}
\section{Related work}
\label{sec:related_work}
\vspace{-2mm}

\inlinesection{Multi-view image generation.}
Early works on multi-view image generation~\citep{watson2022novel, liu2023zero, shi2023mvdream, wang2023imagedream} mainly focus on object-centric scenes without background. Follow-up works~\citep{sargent2024zeronvs, tang2023mvdiffusion,schneider2025worldexplorer} extend multi-view image generators to the scene scale.
CAT3D~\citep{gao2024cat3d} extends a pre-trained image diffusion model for multi-view image generation from any generated or real image input. In a subsequent stage, the generated multi-view images are reconstructed into an explicit 3D representation, e.g., NeRF~\citep{NeRF} or 3DGS~\citep{kerbl20233d}.
Moreover, these methods have been extended to generate dynamic 3D scenes from multiple viewpoints using a space-time diffusion model~\citep{4DiM,wu2024cat4d,kuang2024collaborative,wang20244real}.
Even though these methods demonstrate impressive results, grounding the generations in an explicit 3D representation requires an expensive optimization stage that is not amortized across scenes. Instead of creating multi-view images and then reconstructing using optimization, we are interested in \textit{directly} generating a 3D scene from text or a single image.

\inlinesection{Camera-conditioned video generation.}
There has been significant progress in fine-tuning video diffusion models for 3D camera control.
MotionCtrl~\citep{MotionCtrl} pioneered camera control by conditioning pre-trained video models with camera poses.
Follow-up works~\citep{CameraCtrl, xu2024camco, bahmani2024ac3d, bahmani2024vd3d} represent cameras as Plücker coordinates for pixel-wise conditioning. Another line of work~\citep{MotionMaster,hou2024training,zhang2024recapture} investigates training-free camera control of video diffusion models.
Recently, SynCamMaster~\citep{bai2024syncammaster} and ReCamMaster~\citep{bai2025recammaster} demonstrated impressive camera-controlled video synthesis using large-scale synthetic training data.
While these works achieve compelling results, their output is not grounded in 3D. Our work is orthogonal to this line of work, as we tackle the task of feed-forward 3D generation from camera-controlled video models, allowing novel viewpoint rendering \textit{after} generation. Concretely, we build upon GEN3C~\citep{ren2025gen3c}, a recent camera-controlled video diffusion model, and use it as a teacher within a self-distillation framework for 3D reconstruction.

\inlinesection{Feed-forward 3D models.}
Early work on feed-forward 3D reconstruction mainly focused on object-centric scenes without background. These methods directly predict a NeRF~\citep{hong2023lrm,li2023instant3d} or 3DGS~\citep{LGM,szymanowicz2023splatter} from either text or an image.
More recent works~\citep{charatan2024pixelsplat, zhang2024gs, szymanowicz2024flash3d,ren2024scube} focus on scene-level 3D reconstruction, typically regressing per-pixel 3D Gaussians from one or more input images.
However, most scene-level 3D reconstruction methods are limited to training distribution scenes, e.g., RealEstate10K~\citep{RealEstate10k}, with limited generalizability to generated scenes.
Recent methods extend feed-forward 3D reconstruction models to generate scenes.
Bolt3D~\citep{szymanowicz2025bolt3d} trains a pointmap~\citep{dust3r} autoencoder to generate pointmaps along the multi-view images, used for feed-forward 3D reconstruction. 
Closely related to our work, Wonderland~\citep{liang2024wonderland} uses a camera-controlled video model to generate 3D Gaussians with a feed-forward network.
In contrast to these works, we use a camera-controlled video model as a teacher within a self-distillation framework to train our student 3D model \textit{without requiring real-world multi-view data}. Importantly, we show a simple extension of our framework for feed-forward 4D scene generation, an unexplored task to date.
\vspace{-2mm}
\section{Self-Distillation using Video Diffusion Models}
\label{sec:method_data_engine}
\vspace{-2mm}
Our key idea is to distill the implicit 3D knowledge embedded in video diffusion models into an explicit 3DGS decoder capable of producing high-quality 3D representations.
To this end, we build a teacher–student framework in which the video diffusion model (teacher) generates RGB videos that supervise the 3DGS decoder (student) that operates in the same latent space, as illustrated in Fig.~\ref{fig:self_distillation}.
In this section, we detail our video diffusion model backbone (Sec.~\ref{sec:method_camera_control}) and its role in self-distillation for the 3DGS decoder (Sec.~\ref{sec:method_self_distillation}). 
We discuss the design choice for the 3DGS decoder in Sec.~\ref{sec:method_recon} and outline minimal changes required to adapt the pipeline to dynamic 3D scenes in Sec.~\ref{sec:method_4d}.

\subsection{Background: Camera-Controlled and 3D-Consistent Video Diffusion}
\label{sec:method_camera_control}

Since we supervise our 3DGS decoder only with the camera-controlled video diffusion model, we heavily rely on its 3D consistency. Due to this, we build our approach on GEN3C~\citep{ren2025gen3c}, a recent camera-conditioned video diffusion model. In the following we provide an overview of the key design choices made in GEN3C.

\inlinesection{Spatiotemporal 3D cache.} To improve video consistency and camera control precision, GEN3C constructs a spatiotemporal 3D cache $\{\Point^{\Time, \View}\}$ from input image(s) or videos, where each $\Point^{\Time, \View}$ is a colored point cloud obtained by unprojecting the depth estimation~\citep{wang2024moge} of an RGB image captured from camera viewpoint $\View$ at time $\Time$. The cache is organized as an $\videoLength \times \nView$ array, where $\videoLength$ denotes the number of frames (temporal length) and $\nView$ denotes the number of camera views. 

\inlinesection{Rendering and structured guidance.} To leverage this cache, GEN3C renders each point cloud from arbitrary given camera poses, producing RGB images $\RGBVideo^{\Time, \View}$ and disocclusion masks $\MaskVideo^{\Time, \View}$ via $(\RGBVideo^{\Time, \View}, \MaskVideo^{\Time, \View}) = \mathcal{R}(\Point^{\Time, \View}, \Camera^\Time)$, where $\mathcal{R}$ denotes the rendering function that projects the 3D point cloud $\mathbf{P}^{t,v}$ onto the 2D camera plane according to the camera pose $\mathbf{C}^t$ at time $t$. 
Given a sequence of camera poses $\Camera= (\Camera^1, \dots, \Camera^\videoLength)$, rendering all cache elements produces $\nView$ videos of length $\videoLength$, which can be stacked into image sequences $\RGBVideo^v \in \mathbb{R}^{\videoLength \times 3 \times H \times W}$ and mask sequences $\MaskVideo^\View \in \mathbb{R}^{\videoLength \times 1 \times H \times W}$ for each view $\View$.
The disocclusion masks indicate areas that the video diffusion model should fill in. 
These renderings serve as structured visual guidance for subsequent video generation.

\inlinesection{Video variational autoencoder (VAE).} 
In diffusion-based video generation models, a video variational autoencoder (VAE)~\citep{kingma2013auto,LatentDiffusion} is commonly employed to compress videos into a lower-dimensional latent space for efficient training and inference. 
Given a RGB video $\RGBVideo\in\mathbb{R}^{\videoLength \times 3 \times H \times W}$, a pre-trained VAE encoder $\vaeEncoder$ will encode the video into a latent space, i.e. $\latent= \vaeEncoder(\RGBVideo) \in\mathbb{R}^{\videoLength^{\prime} \times C \times h \times w}$.
The training and inference of the diffusion model are then performed in this latent space. 
The final video $\hat{\RGBVideo} = \rgbDecoder(\latent)$ is decoded with a pre-trained VAE decoder $\rgbDecoder$. In this paper, we adopt the pre-trained GEN3C~\citep{ren2025gen3c,agarwal2025cosmos} model. Specifically, the latent channel dimension is $C = 16$, the temporal dimension is $\videoLength^{\prime} = (\videoLength-1)/\tau + 1$, and the spatial dimensions are $h = H/\sigma$ and $w = W/\sigma$, where the temporal compression factor is $\tau = 8$ and the spatial compression factor is $\sigma = 8$.

\subsection{Self-Distillation}
\label{sec:method_self_distillation}

We train the 3DGS decoder $\gsDecoder$ as a student under a teacher–student paradigm, where a camera-controlled video diffusion model $\videoDiffusionModel$ serves as the teacher. For diverse supervision, we curate a large-scale set of diverse text prompts with large language models~\citep{chatgpt,Qwen-VL}, generate images $\RGBImage$ with an image diffusion model~\citep{flux2024}, and expand into multi-view sequences using GEN3C~\citep{ren2025gen3c}.

\inlinesection{Teacher–student setup.}
Given an input image $\RGBImage$ and a sampled camera trajectory $\{\Camera^\Time\}_{\Time=1}^\videoLength$, the video diffusion model $\videoDiffusionModel$ generates a denoised video latent $\latent = \videoDiffusionModel(\RGBImage, \{\Camera^\Time\}_{\Time=1}^\videoLength)$. The latent $\latent$ is decoded along two branches: the pre-trained RGB decoder $\rgbDecoder$ produces video frames $\RGBVideo_{\rgbDecoder} = \rgbDecoder(\latent)$, while the 3DGS decoder $\gsDecoder$  outputs explicit 3D Gaussians $\gaussians$. Rendered views from 3DGS \textit{(student)} are defined as $\RGBVideo_{\gsDecoder} = \mathrm{Render}(\gaussians, \{\Camera^\Time\}_{\Time=1}^\videoLength)$ and are supervised to match RGB frames $\RGBVideo_{\rgbDecoder}$ \textit{(teacher)}, forming the self-distillation loop.

\begin{wrapfigure}{r}{0.21\linewidth}
\vspace{-2.5em}
\begin{center}
\includegraphics[width=\linewidth]{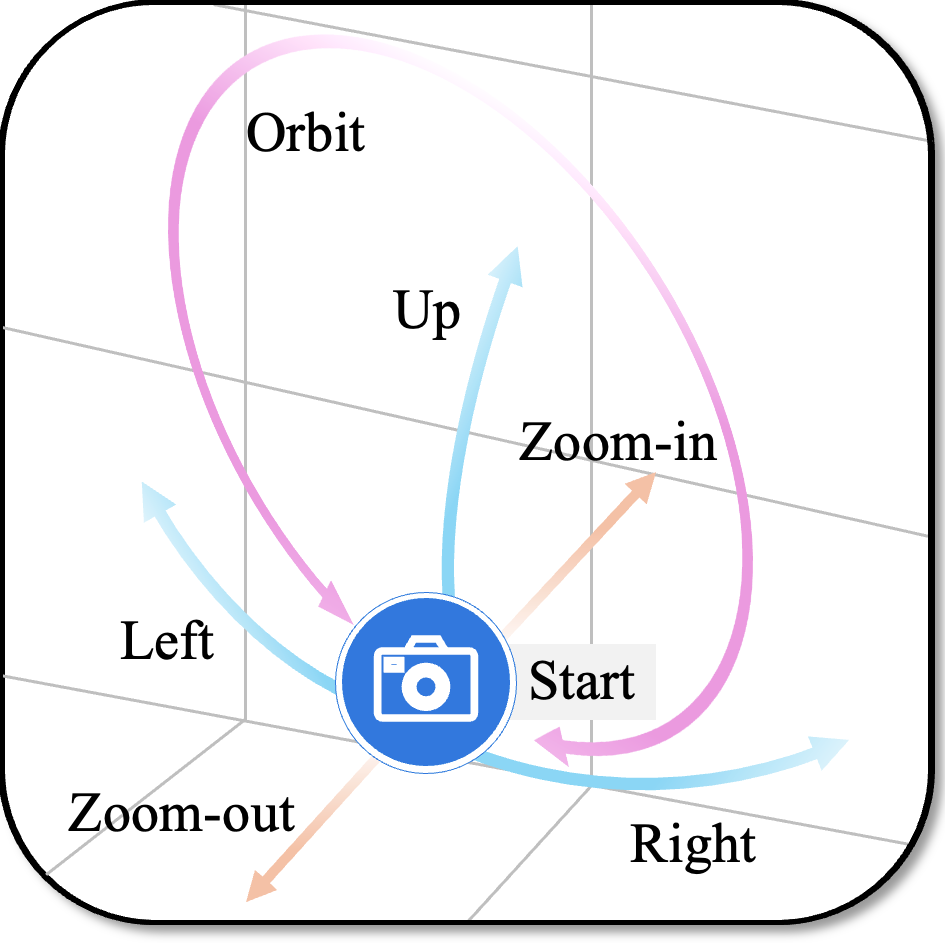}
\caption{{Sampled six camera trajectories to maximize view coverage.}}
\label{fig:sdg_traj}
\vspace{-1.5em}
\end{center}
\end{wrapfigure}
\inlinesection{Multi-trajectory supervision.}
In our paper, to enlarge viewpoint coverage for the input image, we sample $\nView=6$ camera trajectories per input image, as shown in Fig.~\ref{fig:sdg_traj}. For each trajectory $\View$, we construct a spatiotemporal cache $\{\Point^{\Time, \View}\}_{\Time=1}^\videoLength$ along with  $\videoLength=121$ camera poses $\{\Camera^{\Time,\View}\}_{\Time=1}^\videoLength$. Passing these caches through video diffusion model $\videoDiffusionModel$ yields latents $\latent^{v}$, which the teacher decodes into RGB frames $\RGBVideo_{\rgbDecoder}^{\View} = \rgbDecoder(\latent^{v})$. The 3DGS decoder $\gsDecoder$ learns to fuse these multiple $\latent^{v}$ into coherent Gaussians $\gaussians$, while filling disoccluded regions. Through this self-distillation design, the 3DGS decoder $\gsDecoder$ is trained entirely from synthetic supervision provided by $\videoDiffusionModel$. Operating in latent space enables efficient aggregation of multiple trajectories, while the explicit Gaussian output $\gaussians$ enforces geometric consistency and supports downstream simulation and real-time rendering.

\begin{figure}[t]
\centering
\includegraphics[width=\textwidth]{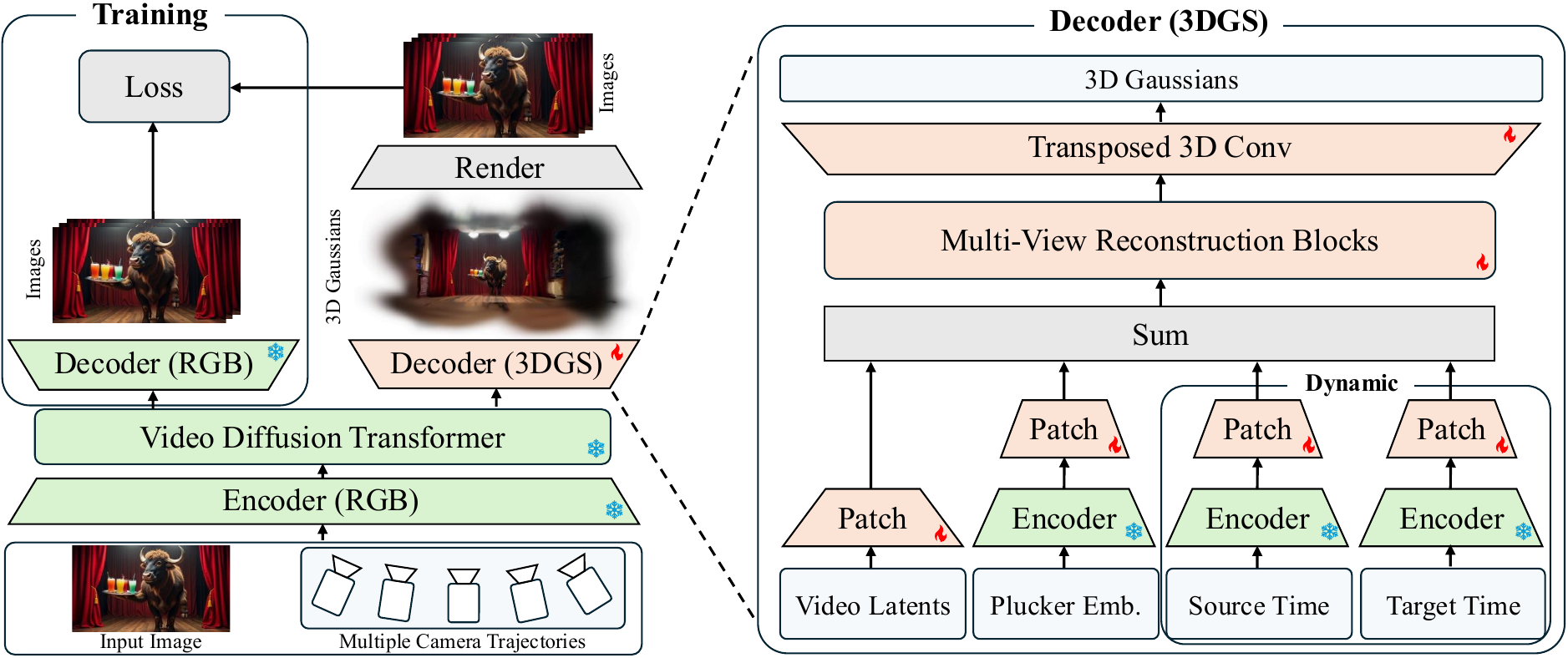}
\caption{\textbf{3D Generative Reconstruction Framework.} Our pipeline builds upon a camera-controlled video diffusion model~\citep{ren2025gen3c} pre-trained on large scale data. We train a 3D Gaussian Splatting (3DGS) decoder by aligning the 2D image renderings of generated 3DGS scenes with the RGB-decoded generations of the pre-trained video model. We only train the 3DGS decoder while freezing the pre-trained autoencoder and diffusion model.
At inference time we directly use the 3DGS decoder, without requiring the RGB decoder anymore.
Time conditioning within the 3DGS decoder allows us to easily extend our approach from static to dynamic 3D scene generation.
}
\label{fig:method}
\end{figure}

\vspace{-2mm}
\section{Feed-Forward Reconstruction from Multi-View Video Latents}
\label{sec:method_recon}
\vspace{-2mm}
Our feed-forward 3DGS decoder $\mathcal{D}_s$ is designed to transform the synthesized multi-view latents generated by our video model $\mathcal{V}$ into an explicit 3D representation that can be rendered from arbitrary viewpoints. We choose 3D Gaussian Splatting (3DGS)~\citep{kerbl20233d} due to its explicit representation and fast rendering speed. We visualize our detailed 3DGS decoder in Fig.~\ref{fig:method} (right).

\subsection{3DGS Decoder}
\label{sec:decoder}

\inlinesection{Scaling Latent-Based 3D Reconstruction.}
Previous feed-forward reconstruction frameworks generate 3D Gaussians for each pixel, and thus, are limited by the number and resolution of input views they can handle. For example, GS-LRM~\citep{zhang2024gs} operates on 2–4 images at $512 \times 512$ resolution, while AnySplat~\citep{jiang2025anysplat} is trained with 24 images at $448 \times 448$ resolution. 
Our video model synthesizes $\nView \times \videoLength = 6 \times 121 = 726$ input views at $704 \times 1280$ resolution—far beyond the capacity of existing methods, which exceeds GPU memory limit.
The bottleneck lies in the attention mechanism applied to visual tokens, whose memory and compute requirements grow with the number of pixels. To overcome this limitation, we avoid scaling in pixel space and instead operate directly in the compressed video latent space produced by our camera-controlled video diffusion model.
The multi-view video latents are denoted as $\allLatent \in \mathbb{R}^{\nView \times \videoLength^{\prime} \times C \times h \times w}$.

\inlinesection{Architecture.}
The 3DGS decoder $\gsDecoder$ first maps its inputs into the hidden dimension of the main reconstruction blocks and outputs per-pixel 3D Gaussian features $\gaussians \in \mathbb{R}^{V \times L \times H \times W \times 14}$. Specifically, the inputs are the multi-view video latents $\allLatent$ and encoded Plücker embeddings $\plucker$, both with latent dimensions $\allLatent, \plucker \in \mathbb{R}^{\nView \times \videoLength^{\prime} \times C \times h \times w}$. 
Each input component is first patchified to match the hidden dimension of the main reconstruction blocks. We do not require any additional visual encoder—only a spatial $2 \times 2$ patchification layer~\citep{ViT} to transform the video latents into flattened tokens for the reconstruction network. The sum of both inputs is processed through the reconstruction blocks. We follow the design introduced in Long-LRM~\citep{ziwen2024long}, i.e., one block consists of one Transformer~\citep{Transformer} layer followed by seven Mamba-2~\citep{dao2024transformers} layers. We repeat the block twice to get 16 layers with 512 hidden dimensions. Finally, a transposed 3D convolution maps the hidden representation to 14 Gaussian channels: 3D position $(x,y,z)$, scale $(s_x,s_y,s_z)$, rotation quaternion $(q_w,q_x,q_y,q_z)$, opacity $\alpha$, and RGB $(r,g,b)$. Formally, the static decoder is expressed as $\gaussians = \gsDecoder(\allLatent, \plucker)$.

\inlinesection{Plücker embeddings.}
Raw Plücker embeddings are first computed from camera poses $\{\Camera^{\Time,\View}\}$ and intrinsics as $\plucker^{\text{raw}} \in \mathbb{R}^{\nView \times \videoLength \times 6 \times H \times W}$. We reuse the RGB encoder $\vaeEncoder$ from the pre-trained VAE to encode  Plücker embeddings. Specifically, we separately encode the 3-dimensional ray directions and the 3-dimensional cross product of ray directions and origins using  $\vaeEncoder$, and concatenate latent along the channel dimension to obtain $\mathbf{E}^{\text{enc}} \in \mathbb{R}^{\nView \times \videoLength^{\prime} \times 2C \times h \times w}$. A lightweight MLP maps $\mathbf{E}^{\text{enc}}$ to $\mathbf{E} \in \mathbb{R}^{\nView \times \videoLength^{\prime} \times C \times h \times w}$, reducing the channel dimension by a factor of 2 to match video latent $\allLatent$.

\subsection{Loss function}
\label{sec:loss}

\inlinesection{Image-based supervision.} We supervise our 3DGS decoder with an image-based reconstruction loss. The reconstruction loss is split into a Mean Squared Error (MSE) loss $\mathcal{L}_{mse}$ and an LPIPS loss $\mathcal{L}_{lpips}$~\citep{zhang2018unreasonable} using VGG~\citep{simonyan2014very} as a feature extractor.

\inlinesection{Depth supervision.} We observed that the decoder trained with only RGB loss often produces flattened geometry. To address this, we additionally supervise the rendered depth maps using the consistent video depth estimated by an off-the-shelf system ViPE~\citep{huang2025vipe}, and the scale-invariant depth loss $\mathcal{L}_{depth}$ from Long-LRM~\citep{ziwen2024long}.

\inlinesection{Opacity-based pruning.} Similar to Long-LRM~\citep{ziwen2024long}, we use an L1 regularization on the opacity $\mathcal{L}_{opacity}$ and remove the Gaussians with the lowest 80\% opacity.

\inlinesection{Total loss.}
Our total loss is computed as the weighted sum of all losses with weight factors $\lambda_{i}$
\begin{equation}
\mathcal{L} = 
\lambda_{mse} \, \mathcal{L}_{mse} +
\lambda_{lpips} \, \mathcal{L}_{lpips} +
\lambda_{depth} \, \mathcal{L}_{depth} +
\lambda_{opacity} \, \mathcal{L}_{opacity}
\end{equation}

where we set $\lambda_{mse}=1.0$, $\lambda_{lpips}=0.5$, $\lambda_{depth}=0.05$, and $\lambda_{opacity}=0.1$.

\vspace{-2mm}
\section{Extension to Dynamic 3D Scenes}
\label{sec:method_4d}
\vspace{-2mm}

Our approach can be extended to handle dynamic 3D scenes with minimal changes, outlined below.

\inlinesection{Self-distillation for dynamic scenes.}
Our dynamic 3D setup closely follows the design of the static 3D counterpart. Instead of a single image, the video model takes a single video of length $L$ with corresponding camera poses as input and generates multi-view video latents capturing the same underlying motion. 
For video inputs, we follow a protocol similar to the static case: sampling diverse text prompts with large language models, then generating videos using video diffusion models~\citep{agarwal2025cosmos,wan2025wan}. The generated videos are then annotated with camera poses and depth maps using ViPE~\citep{huang2025vipe}.

\begin{figure}[t]
\centering
\includegraphics[width=\textwidth]{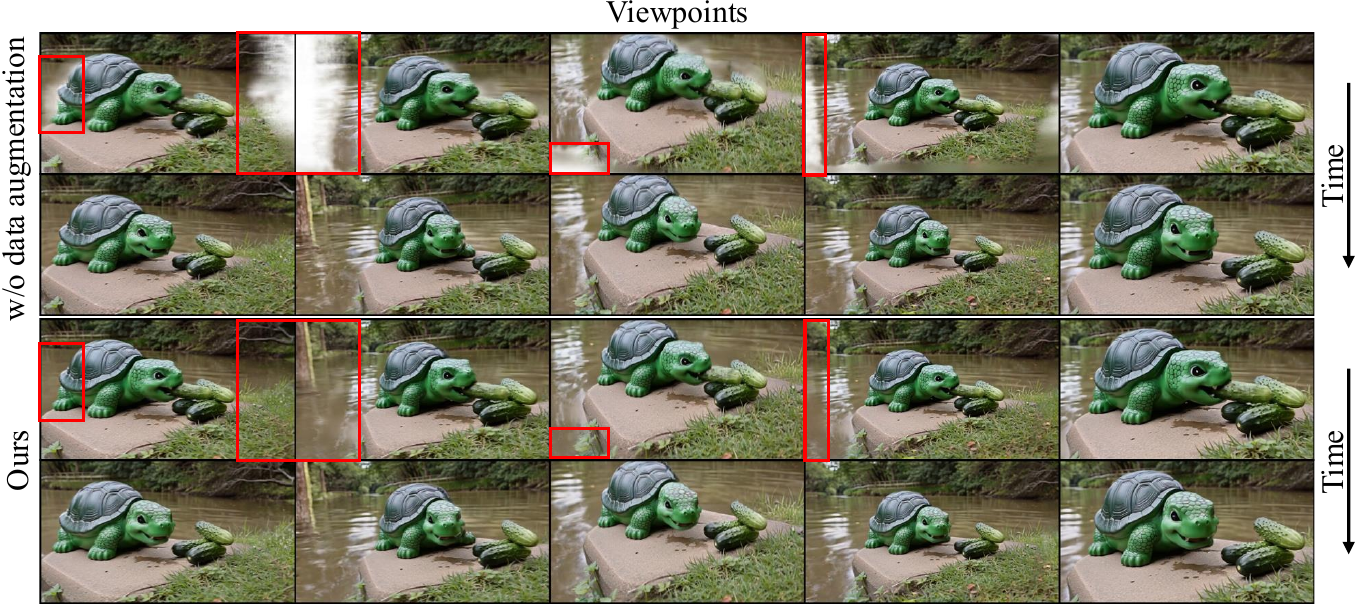}
\caption{\textbf{Dynamic data augmentation.} When naively supervising the time-aware 3DGS decoder, we observe artifacts in the generated 3D Gaussians. Specifically, early timesteps from extreme poses exhibit low opacity in regions not covered by the supervision signal. To address this, we augment the supervision data with a motion-reversed video, ensuring that each timestep is observed from the full spatial coverage, thereby preventing low opacity artifacts in the early timesteps.}
\label{fig:ablations_dynamic}
%\vspace{-0.1cm}
\end{figure}

\inlinesection{Dynamic 3DGS decoder architecture.}
We follow the bullet-time design of \cite{liang2024btimer} and design our dynamic  3DGS decoder $\dynamicgsDecoder$ to generate time-dependent 3D Gaussians for all the video frames. Specifically, the dynamic 3DGS decoder $\dynamicgsDecoder$ closely follows the architecture of $\gsDecoder$, except that we augment the $\dynamicgsDecoder$ input with the encoded source and target time embeddings $\mathbf{T}^{\text{src}}, \mathbf{T}^{\text{tgt}} \in \mathbb{R}^{\nView \times \videoLength^{\prime} \times C \times h \times w }$. The unencoded (raw) source times are assigned to each input frame, forming $\mathbf{T}^{\text{src,raw}} \in \mathbb{R}^{\nView \times \videoLength \times 1 \times H \times W}$, and the target time is $\mathbf{T}^{\text{tgt,raw}} \in \mathbb{R}^{\nView \times 1 \times 1 \times H \times W}$. Both are spatially repeated and augmented with a 2-dimensional sinusoidal embedding to expand the channel dimension to 3, then encoded with the RGB encoder $\vaeEncoder$ to latent dimensions $\mathbf{T}^{\text{src}}, \mathbf{T}^{\text{tgt}} \in \mathbb{R}^{\nView \times \videoLength^{\prime} \times C \times h \times w}$. The target time is repeated along the latent temporal dimension to match $\videoLength^{\prime}$. During training, a target time step is randomly sampled, and the corresponding 3D Gaussians are decoded and supervised across all trajectories generated with dynamic data augmentation. We fine-tune $\mathcal{D}_d$ from a pre-trained $\mathcal{D}_s$ with the patchification layers of $\mathbf{T}^{\text{src}}$ and $ \mathbf{T}^{\text{tgt}}$ initialized to zeros. $\mathbf{T}^{\text{src}}$ and $ \mathbf{T}^{\text{tgt}}$ are directly added to the sum of $\allLatent$ and $\plucker$. Formally, the dynamic decoder is defined as $\gaussians= \mathcal{D}_d(\allLatent, \plucker, \mathbf{T}^{\text{src}}, \mathbf{T}^{\text{tgt}})$.

\inlinesection{Dynamic data augmentation.}
In the static 3D scenes, where time is not a factor, frames from different timesteps can all be used to supervise the 3DGS renderings. In contrast, for dynamic 3D scenes, the 3DGS changes over time, so only frames from the corresponding timestamp are valid for supervision. This restriction can lead to a trivial solution, where the 3DGS at a given timestep simply ignores information from frames at other timesteps (see Fig.~\ref{fig:ablations_dynamic}). 

To mitigate this issue, we introduce a dynamic data augmentation strategy that balances supervision across timesteps. Early frames are naturally associated with viewpoints close to the input image, whereas later frames correspond to more distant viewpoints. To counter this imbalance, we augment the training data with paired supervision views for each timestep — one near and one far. 
Concretely, we reverse the input video in frame order and feed it into the video model $\mathcal{V}$, which produces six additional multi-view sequences. After reversing these sequences back to the original motion order, we obtain six trajectories that progress inward (from far viewpoints toward the input). Combined with the original six outward trajectories, this yields 12 supervision views per timestep. Importantly, this augmentation is applied only during training: the reversed trajectories act purely as extra supervision signals to prevent collapse from pruning artifacts and are not required at inference time.

\vspace{-2mm}
\section{Experiments}
\label{sec:experiments}
\vspace{-2mm}

\subsection{Experimental Setup}
\label{sec:exp_setup}
\inlinesection{Datasets.}
To train our model, we do not use any existing multi-view datasets; instead, we construct our own, which we call the \methodname~dataset, relying solely on our video model to supervise the 3DGS decoder. The 3D reconstruction setup uses 59,031 images, while the 4D setup has 7,378 videos. All data are from diverse text prompts, spanning scenarios such as indoor and outdoor environments, humans, animals, and both realistic and imaginative content. We synthesize six camera trajectories for each image (3D) or video (4D), yielding 354,186 videos for 3D and 44,268 videos for 4D.

\inlinesection{Baselines.}
We compare our method with the state-of-the-art approaches for generative single-image-to-3D reconstruction, i.e., ZeroNVS~\citep{sargent2024zeronvs}, ViewCrafter~\citep{yu2024viewcrafter}, Wonderland~\citep{liang2024wonderland}, and Bolt3D~\citep{szymanowicz2025bolt3d}. 
Note that \textit{no source code is available} to evaluate these methods on our out-of-distribution evaluation set, hence we mainly rely on reported quantitative comparisons from the papers.

\inlinesection{Evaluation.}
We follow previous works to evaluate our model on the task of single-image to 3D using RealEstate10K~\citep{RealEstate10k}, DL3DV~\citep{ling2024dl3dv}, and Tanks and Temples~\citep{Knapitsch2017}. We follow the evaluation protocol outlined in Wonderland~\citep{liang2024wonderland} and Bolt3D~\citep{szymanowicz2025bolt3d}. We evaluate the performance using standard reconstruction metrics, i.e., PSNR, SSIM, and LPIPS.

\begin{table}[t!]
\small
\centering
\setlength{\tabcolsep}{4pt}
\begin{tabular}{l ccc ccc ccc}
\toprule
\multirow{2}{*}{\textnormal{Method}} 
& \multicolumn{3}{c}{\textnormal{RealEstate10K}} 
& \multicolumn{3}{c}{\textnormal{DL3DV}} 
& \multicolumn{3}{c}{\textnormal{Tanks-and-Temples}} \\
\cmidrule(lr){2-4} \cmidrule(lr){5-7} \cmidrule(lr){8-10}
& PSNR~$\uparrow$ & SSIM~$\uparrow$ & LPIPS~$\downarrow$ 
& PSNR~$\uparrow$ & SSIM~$\uparrow$ & LPIPS~$\downarrow$ 
& PSNR~$\uparrow$ & SSIM~$\uparrow$ & LPIPS~$\downarrow$ \\
\midrule
ZeroNVS     & 13.01 & 0.378 & 0.448 & 13.35 & 0.339 & 0.465 & 12.94 & 0.325 & 0.470 \\
ViewCrafter & 16.84 & 0.514 & 0.341 & 15.53 & 0.525 & 0.352 & 14.93 & 0.483 & 0.384 \\
Wonderland  & \textnormal{17.15} & \textnormal{0.550} & \textnormal{0.292} & \textnormal{16.64} & \textnormal{0.574} & \textnormal{0.325} & \textnormal{15.90} & \textnormal{0.510} & \textnormal{0.344} \\
Bolt3D      & \textnormal{21.54} & \textnormal{0.747} & \textnormal{0.234} & \textnormal{-} & \textnormal{-} & \textnormal{-} & \textnormal{-} & \textnormal{-} & \textnormal{-} \\
\midrule
Ours & \textbf{21.79} & \textbf{0.752} & \textbf{0.219} & \textbf{20.09} & \textbf{0.583} & \textbf{0.313} & \textbf{19.24} & \textbf{0.570} & \textbf{0.336} \\
\bottomrule
\end{tabular}
\caption{\textbf{State-of-the-art comparisons.} We compare our method with previous works for single image to 3D generation using RealEstate10K, DL3DV, and Tanks-and-Temples datasets.
}
\label{tab:sota_comparison}
\vspace{-0.5em}
\end{table}

\begin{figure}[t!]
\centering
\includegraphics[width=1.0\textwidth]{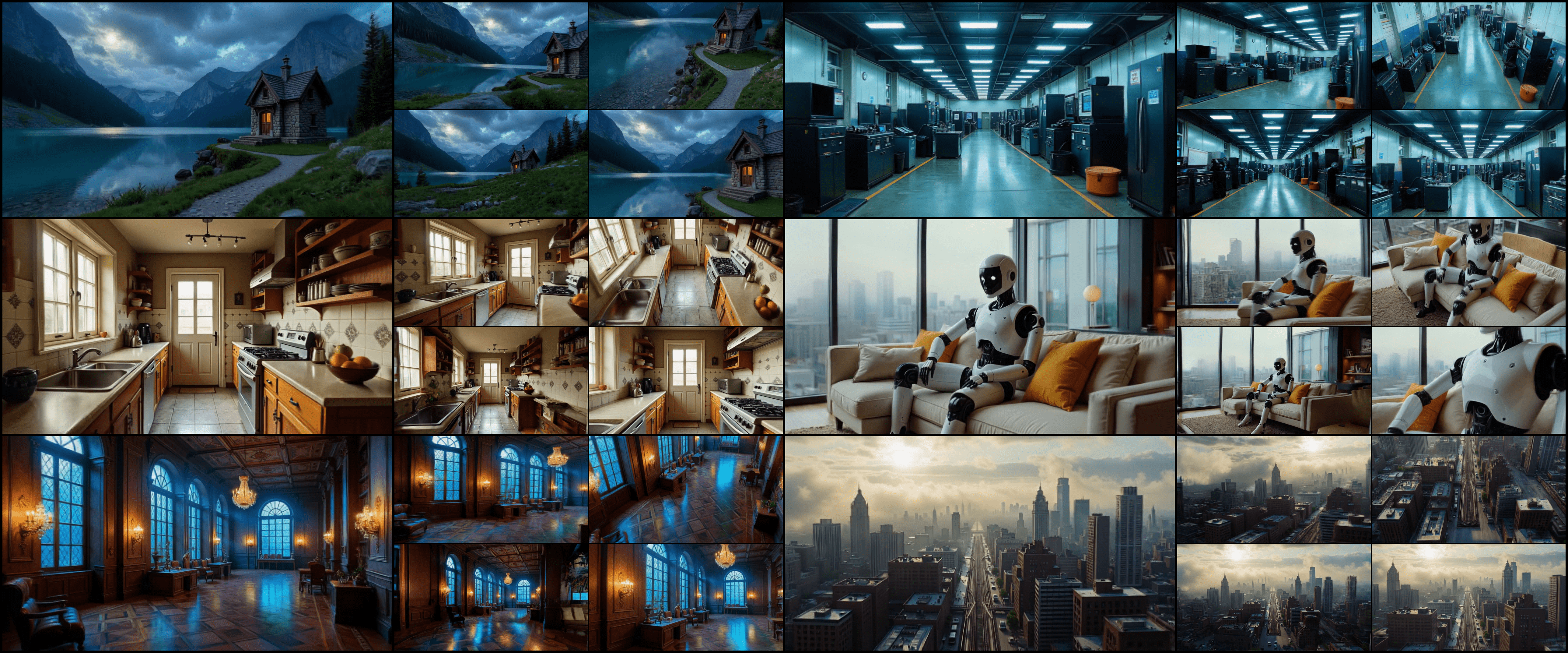}
\vspace{-0.7em}
\caption{\textbf{Image-to-3DGS Generation.} We visualize five views from generated 3DGS scenes.
}
\label{fig:main_results}
\vspace{-1em}
\end{figure}

\subsection{Main Results}
\label{sec:main_results}

\inlinesection{Quantitative results.}
We show quantitative evaluations for single image-to-3D in Tab.~\ref{tab:sota_comparison}. Our method outperforms previous works on all benchmarks across all metrics.
Hence, the main improvements to further boost the quality lie in the development of stronger video generative models, as our reconstructions will directly benefit from them. We provide additional quantitative comparisons on the \methodname~dataset in Appendix~\ref{sec:evaluation_details}.

\inlinesection{Qualitative results.}
We visualize novel view renderings from our generated 3DGS scenes in Fig.~\ref{fig:teaser} and Fig.~\ref{fig:main_results}, but strongly recommend the reader to our \ul{supplementary webpage} for video results.
Our method generates high-quality novel view content for 3D/4D scenes in unseen regions while maintaining consistency with the input image/video. We visualize qualitative comparisons on the \methodname~dataset in Appendix~\ref{sec:evaluation_details}.

\begin{figure}[t]
\centering
\includegraphics[width=\textwidth]{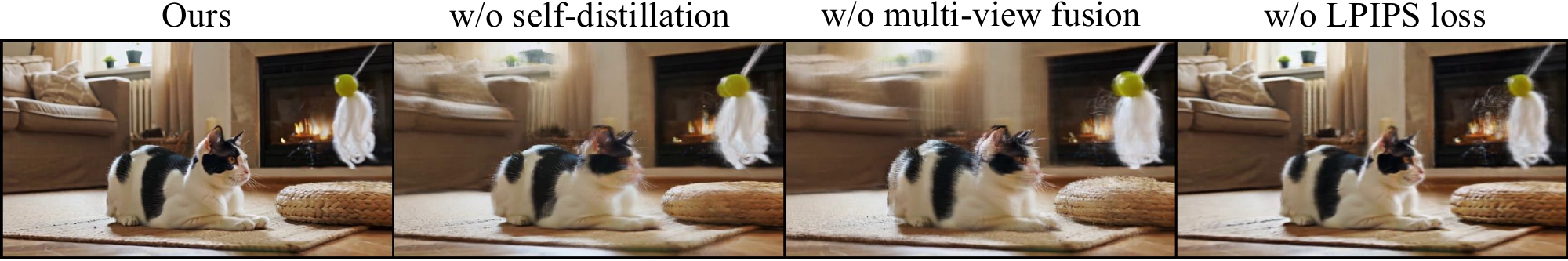}
\caption{\textbf{Ablations.} We visualize ablation results by rendering the same extreme novel viewpoint after image-to-3DGS generation; depth visualizations and ablations are provided in Appendix~\ref {sec:depth_visualization}. 
}
\label{fig:ablations}
\vspace{-0.0cm}
\end{figure}

\subsection{Ablations}
\label{sec:ablations}

In Tab.~\ref{tab:ablations}, we motivate our design choices by ablating key components on out-of-distribution diverse prompts.
We compare 3DGS renderings with videos generated by our camera-controlled video diffusion model.

% \inlinesection{Self-distillation.}
\inlinesection{Real data only.}
Instead of using self-distillation, we train a model using only real multi-view datasets, i.e., RealEstate10K~\citep{RealEstate10k} and DL3DV~\citep{ling2024dl3dv}. These datasets are commonly used in previous works, but lead to limited generalizability to out-of-distribution scenes.

\begin{wraptable}{r}{0.5\linewidth}
\vspace{-1em}
\centering
\small
\setlength{\tabcolsep}{4pt}
\begin{tabular}{lccc}
\toprule
Method & PSNR~$\uparrow$ & SSIM~$\uparrow$ & LPIPS~$\downarrow$ \\
\midrule
Ours & 24.77 & 0.837 & 0.224 \\
\midrule
\multicolumn{4}{l}{\textbf{Data}} \\
% w/o self-distill.      & 19.08 & 0.659 & 0.413 \\
% w/o self-distill. only & 24.74 & 0.823 & 0.236 \\
real data only      & 19.08 & 0.659 & 0.413 \\
{self-distill. + real data} & 24.74 & 0.823 & 0.236 \\
\midrule
\multicolumn{4}{l}{\textbf{Loss}} \\
w/o depth loss         & 24.31 & 0.811 & 0.247 \\
w/o opacity pruning    & 24.55 & 0.820 & 0.237 \\
w/o LPIPS loss         & 23.74 & 0.766 & 0.370 \\
\midrule
\multicolumn{4}{l}{\textbf{Architecture}} \\
w/o multi-view fusion  & 17.73 & 0.632 & 0.446 \\
w/o Mamba-2            & 24.58 & 0.818 & 0.241 \\
w/o latent 3DGS        &       &  OOM  &       \\
\bottomrule
\end{tabular}
\captionof{table}{\textbf{Ablation study on \methodname\ dataset.}}
\label{tab:ablations}
\vspace{-2em}
\end{wraptable}

% \inlinesection{Self-distillation only.}

\inlinesection{Self-distillation + real data.}
Perhaps surprisingly, joint training with self-distillation and real data does not improve over using only self-distillation without real data. This confirms that self-distillation is diverse and consistent enough to learn a reconstruction model.

\inlinesection{No depth loss.}
Depth loss prevents flat geometry and even slightly improves image-based metrics. We visualize depth maps in Appendix~\ref {sec:depth_visualization}.

\inlinesection{No opacity pruning.}
We keep all Gaussians instead of pruning them by opacity. Learning to prune most of the 3D Gaussians makes the output representation more compact and slightly improves visual quality. Rendering at $H=704$, and $W=1280$ takes 18ms with pruning vs. 30ms without pruning, a $1.67 \times$ speed up.

\inlinesection{No LPIPS loss.}
Adding the LPIPS loss improves the robustness to input inconsistencies and enhances the preservation of high-frequency details.

\inlinesection{No multi-view fusion.}
A naive implementation would be to generate 3D Gaussians for each camera trajectory independently and then fuse the 3D Gaussians into one point cloud. However, we observe significant advantages by learning the multi-view fusion from different trajectories with our reconstruction blocks, where each token attends to the others. 

\inlinesection{No Mamba-2.}
We replace our joint Transformer/Mamba-2 with Transformer-only blocks and observe slightly lower quality.  
One forward pass with $V=6$, $L=121$, $H=704$, and $W=1280$ takes 3213ms with joint blocks in comparison to 20922ms with Transformer-only blocks, a 6.5 $\times$ speedup.

\inlinesection{No latent-based 3DGS.}
Previous works operating in the pixel space, such as BTimer~\citep{liang2024btimer}, are restricted to 12 input frames, while we can take up to 726 input frames. Consequently, operating the 3DGS decoder in the pixel space instead of the latent space leads to out-of-memory.

\vspace{-2mm}
\section{Conclusion}
\label{sec:conclusion}
\vspace{-2mm}

In this work, we propose \methodname, a novel 3D and 4D generation framework relying only on a single image or video input. Instead of collecting multi-view datasets, we introduce camera-controlled video diffusion models as teachers within a self-distillation framework for a student 3DGS decoder. Our 3DGS decoder operates in the latent space of the video model and directly reconstructs 3D Gaussians without any post-processing or optimization. This design enhances generalizability and coverage across diverse scenes, while our dynamic extension demonstrates the feasibility of 4D generation from monocular video.
Currently, the scale and consistency of our generated scenes are bounded by the capacity of our camera-controlled video diffusion model. Hence, further improving the video diffusion model will enable large-scale, consistent scene synthesis. Moreover, investigating the adaptation of auto-regressive techniques~\citep{chen2024diffusion} into our framework for large-scale generation is an interesting direction for future work. 
Lastly, modelling motion and tracking information within the reconstruction network, as in concurrent work~\citep{lin2025movies}, could improve visual motion quality.

\section{Ethics Statement}
Like all generative AI technologies, there are risks of misuse of our method, including producing misleading or synthetic 3D/4D content. While such risks exist, the main goal of this research is to contribute to the fields of simulation, robotics, and embodied AI by providing scalable and controllable tools for data generation. In this sense, we expect our most significant contributions to be made through the following capabilities and insights.

\begin{itemize}
\item \textbf{The ability to enable realistic simulation for embodied agents}: We provide controllable and physically consistent camera motion within generated 3D and 4D environments to support training and evaluating agents that must perceive, navigate, and interact with complex scenes.

\item \textbf{Improved scalability for data engines}: Our method removes the need for multi-view capture and expensive per-scene optimization, and helps produce high-fidelity, customizable scenes at scale, which is particularly suitable for research in robotics, reinforcement learning, and closed-loop simulation.

\item \textbf{Better technical understanding}: As an academic study, this work contributes a foundational case study of how camera control interacts with spatiotemporal generative models, which will benefit research in computer vision, graphics, and embodied AI.
\end{itemize}

While we caution against deceptive or wrongful use of this technology—where misleading and/or unverifiable media could be created with realistic generative outputs—we believe that the adoption of safeguards (e.g., provenance tracking, dataset documentation, and good evaluation practices) will be useful in ensuring that generative models advance science and society in positive ways.

\section{Reproducibility Statement}
\label{sec:reproducibility}
To fully reproduce our results and accelerate future research in this area, we release our training and inference code, model weights, and data for both 3D and 4D generation.

\section{Acknowledgements}
We thank Zian Wang for feedback on the draft.

\bibliography{egbib}
\bibliographystyle{iclr2026_conference}

\newpage
\appendix
\section{Additional Details of Video Diffusion Model}
\label{sec:method_video_details}

This section provides additional information on the GEN3C~\citep{ren2025gen3c} video diffusion model backbone introduced in Sec.~\ref{sec:method_camera_control} and how we further improve its 3D consistency.

\begin{figure}[h]
% \vspace{-0.3cm}
\centering
\includegraphics[width=\textwidth]{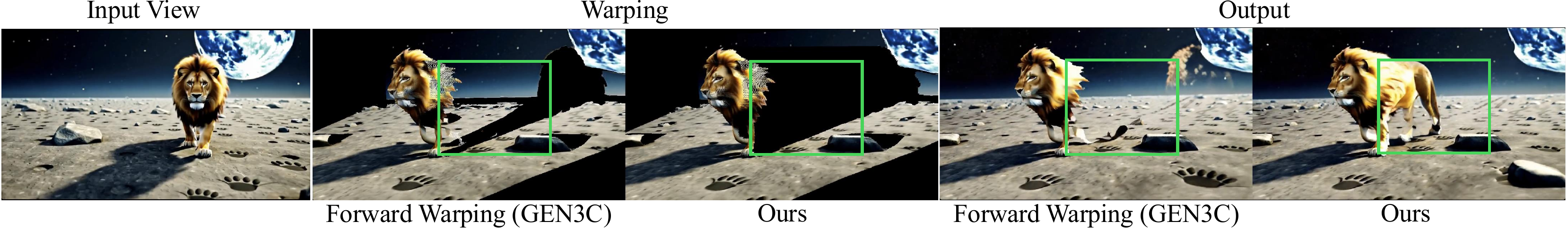}
\caption{\textbf{Rendering function comparison.} Our model builds upon GEN3C~\citep{ren2025gen3c} that uses forward warped images as camera control conditioning. We visualize forward warping (left) vs. our conservative rendering function (right). The highlighted region is incorrectly filled with background pixels in forward warping but properly masked in ours, enabling correct completion.
}
\label{fig:warping_disocclusion}
% \vspace{-0.3cm}
\end{figure}

\inlinesection{Conservative mask refinement.} A key limitation of GEN3C's forward warping is background leakage, where occluded object parts in the source view (e.g., lion body in Fig.~\ref{fig:warping_disocclusion}) become visible in novel viewpoints but are not properly indicated in the disocclusion mask. This leads to incomplete foreground completion by the video diffusion model, as shown in Fig.~\ref{fig:warping_disocclusion} (second and fourth columns).

To address this, we adopt a conservative strategy: since the geometry of occluded regions in $\mathbf{P}^{t,v}$ is unknown, we assume all such areas are occupied and derive refined disocclusion masks accordingly. This allows the video model to reason about potentially visible foreground regions that were missed by standard point-based rendering.

Our approach constructs a triangular mesh from camera-space points $\mathbf{P}_c^{t,v} = \mathbf{C}^t \mathbf{P}^{t,v}$ by connecting spatially adjacent pixels, similar to~\cite{hu2021worldsheet}:
\begin{equation}
\mathcal{M} = \bigcup_{(u,v)} \left\{ \triangle\left(\mathbf{p}_{u,v}, \mathbf{p}_{u+1,v}, \mathbf{p}_{u,v+1}\right), \triangle\left(\mathbf{p}_{u+1,v}, \mathbf{p}_{u+1,v+1}, \mathbf{p}_{u,v+1}\right) \right\}
\end{equation}
where $\mathbf{p}_{u,v}$ denotes the 3D point at pixel $(u,v)$ in $\mathbf{P}_c^{t,v}$. This surface mesh serves as a boundary between visible and invisible regions. For each rendered pixel, we compare the standard point-based depth $\mathbf{D}^{t,v}$ with the mesh-interpolated depth $\mathbf{D}_{\mathcal{M}}^{t,v}$ obtained via ray-surface intersection:
\begin{equation}
\mathbf{M}^{t,v}(u,v) = \begin{cases}
0 & \text{if } \mathbf{D}_{\mathcal{M}}^{t,v}(u,v) < \mathbf{D}^{t,v}(u,v) - \epsilon \\
\mathbf{M}_{\text{orig}}^{t,v}(u,v) & \text{otherwise}
\end{cases},
\end{equation}
where $\mathbf{M}_{\text{orig}}^{t,v}$ denotes the original disocclusion mask from forward warping and $\epsilon$ is a small tolerance factor. When the mesh surface is closer than the original points, we conservatively mask out these regions as potential foreground disocclusions. This produces more reliable guidance for video generation, especially under large viewpoint changes, as shown in Fig.~\ref{fig:warping_disocclusion} (third and fifth columns).

\section{Model Details}
\label{sec:method_details} 

\inlinesection{Progressive training.}
Directly training the 3DGS decoder on $6$ trajectories of $121$ frames at $704 \times 1280$ is expensive, hence we train our network progressively, as shown in Tab.~\ref{tab:training_setup}. Furthermore, we use the final static pre-trained model to initialize our dynamic model. The total training time takes 6 days with 8 NVIDIA A100 80 GB GPUs. We use gsplat~\citep{ye2025gsplat} as the 3DGS implementation.

\inlinesection{Subsampling Gaussians.}
Generating pixel-aligned 3D Gaussians scales linearly with each spatial and temporal resolution. Our generated scenes represent 726 frames at $704 \times 1280$ resolution, leading to 654,213,120 per-pixel Gaussians. Hence, we only generate one 3D Gaussian per $8 \times 8$ spatial neighborhood, reducing the total number of 3D Gaussians by a factor of 64, i.e., 10,222,080. Moreover, our opacity-based pruning further reduces the number to 2,044,416, making the output more compact.

\inlinesection{Joint Transformer-Mamba blocks.}
We use the reconstruction block design introduced in Long-LRM~\citep{ziwen2024long}, i.e., one block consists of one Transformer layer followed by seven Mamba-2 layers. We repeat the block twice to get 16 layers with 512 hidden dimensions. Note that our network is significantly smaller than the one used in Long-LRM~\citep{ziwen2024long} or Wonderland~\citep{liang2024wonderland}, which use 24 layers with 1024 hidden dimensions. 
While pure Transformers are typically more powerful, they are significantly slower in training and inference than Mamba2. Since we trained both configurations within the same training budget, we hypothesize that transformer-only blocks would need considerably more training iterations to fully catch up in rendering quality.

\inlinesection{Camera encoding.}
We use Plücker embeddings for camera encoding, following prior works~\citep{kant2024spad,zhang2024gs}. The embeddings are computed at full spatial and temporal resolution and then projected into the video latent space using the pre-trained RGB encoder. Since the RGB encoder expects 3 input channels but Plücker embeddings are 6-dimensional, we separately encode the 3-dimensional ray directions and the 3-dimensional cross product of ray directions and origins. The resulting latent encodings are then concatenated along the channel dimension.

\inlinesection{Time encoding.}
Similarly, to make the 1-dimensional time input compatible with the 3-channel input expected by the RGB encoder, we concatenate a 2-dimensional sinusoidal embedding to the original time value along the channel dimension. We then replicate these values across the spatial dimensions before encoding the time into the video latent space. This approach allows us to independently encode both the source and target times into the compressed latent representation.

\begin{table}[t]
\centering
\small
\setlength{\tabcolsep}{4pt}
\begin{tabular}{c c c c c c c}
\toprule
Stage & $H \times W$ & $L$ & $V$ & $S$ & $B$ & Steps \\
\midrule
\multicolumn{7}{l}{Static} \\
1 & 176$\times$320  & 17  & 1    & 17   & 4   & 10k  \\
2 & 176$\times$320  & 49  & 1    & 49   & 4   & 2.5k \\
3 & 352$\times$640  & 49  & 1    & 49   & 2   & 2.5k \\
4 & 704$\times$1280 & 49  & 1    & 49   & 1   & 2.5k \\
5 & 704$\times$1280 & 121 & 1    & 9    & 1   & 57.5k \\
6 & 704$\times$1280 & 121 & 1--6 & 9    & 1   & 7k   \\
\midrule
\multicolumn{7}{l}{Dynamic} \\
7 & 704$\times$1280 & 121 & 6    & 12   & 1   & 10k  \\
\bottomrule
\end{tabular}
\caption{\textbf{Progressive training setup.}}
\label{tab:training_setup}
% \vspace{-0.9cm}
\end{table}

\begin{figure}[h]
\centering
\includegraphics[width=\textwidth]{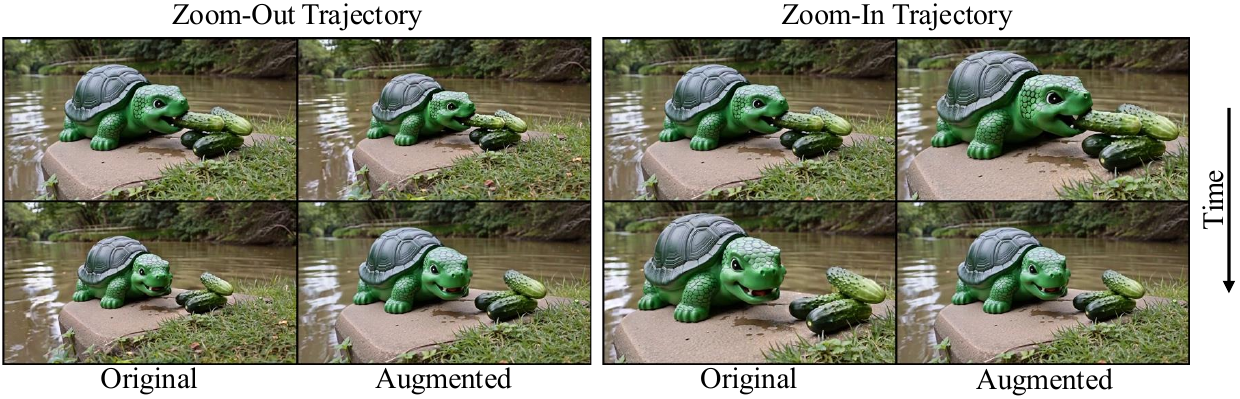}
\caption{\textbf{Dynamic data augmentation videos.} We augment the supervision data with a motion-reversed video, ensuring that each timestep is observed from the full spatial coverage, thereby preventing low opacity artifacts in the early timesteps. We show two example trajectories, i.e., zoom-out and zoom-in, and visualize their corresponding augmented videos. The augmented videos are flipped in their camera motion.}
\label{fig:dynamic_data_augmentation}
\end{figure}

\inlinesection{Dynamic data augmentation.}
We visualize our dynamic data augmentation procedure for two example camera trajectories in Fig.~\ref{fig:dynamic_data_augmentation}. The procedure creates pairs of supervision views that cover the scene from the same motion state but different extreme viewpoints.

\section{Additional Evaluation}
\label{sec:evaluation_details}

\subsection{Additional Baseline details}
\label{sec:baseline_details}

We additionally compare our model against BTimer~\citep{liang2024btimer}, a recent approach for joint 3D and 4D reconstruction from posed images or videos. 
We reached out to the authors of BTimer to conduct experiments for us using the same scenes as \methodname. Since BTimer is purely regression-based, we integrate it with our GEN3C~\citep{ren2025gen3c} video diffusion backbone. Specifically, we use the same camera trajectories as in our model and provide RGB-decoded videos from GEN3C as input to BTimer. BTimer operates in pixel space and requires 12 input images. However, as discussed in the main paper, pixel-space models such as BTimer run out of memory when directly using the 726 high-resolution frames generated by GEN3C. To address this, we uniformly subsample GEN3C frames. For static evaluations, we select 12 frames that evenly cover the available viewpoints. For dynamic evaluations, we sample 11 frames to span the viewpoint range and additionally include the bullet-time frame corresponding to the motion being reconstructed. This ensures that the target motion state is always present as a reference. We denote this integrated baseline as \btimerbaseline, i.e., BTimer combined with GEN3C.

\subsection{Additional Static 3D Evaluation}
\label{sec:evaluation_3d}
\begin{figure}[t]
\centering
\includegraphics[width=\textwidth]{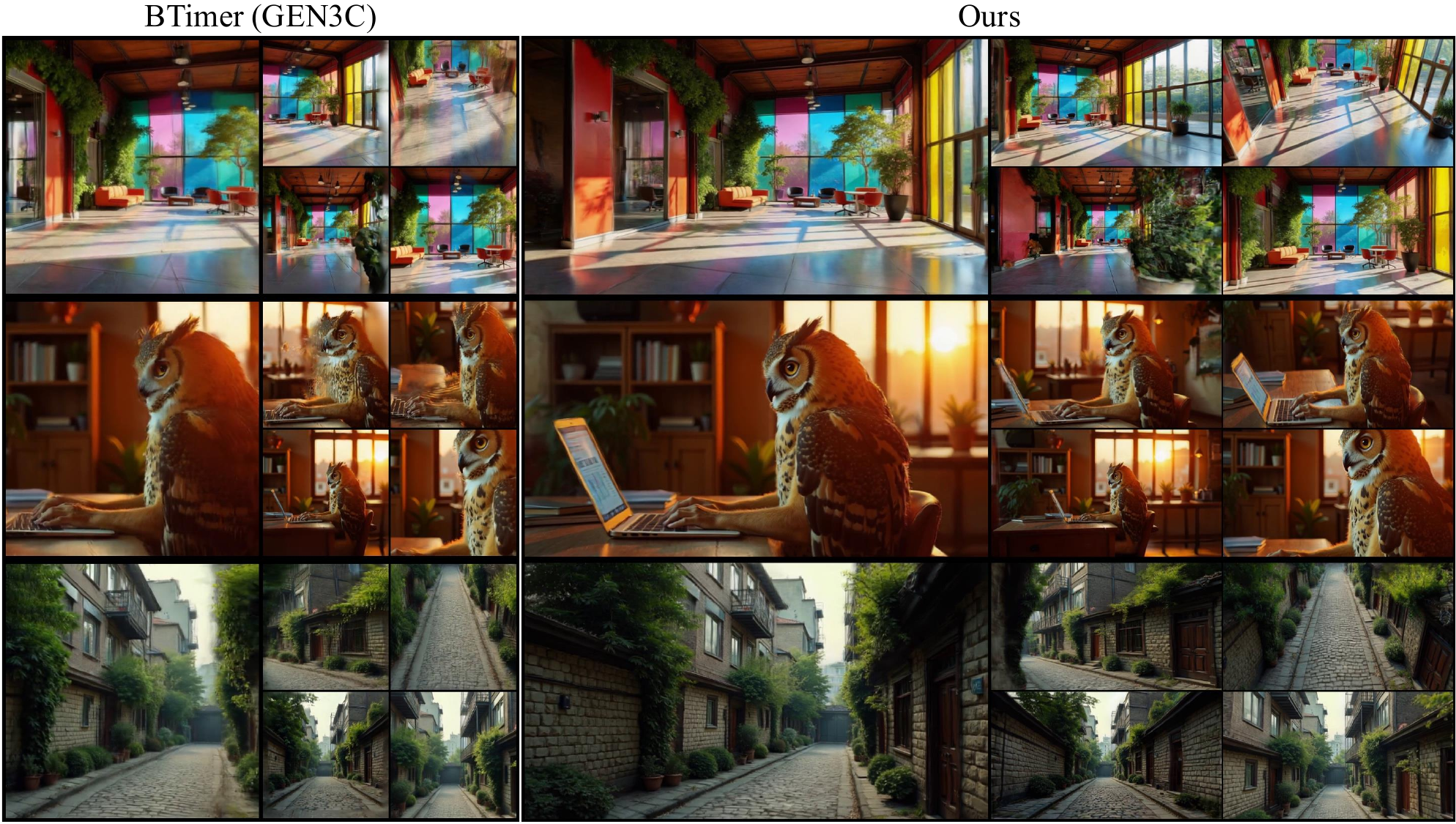}
\caption{\textbf{Image-to-3DGS Comparison.}
We compare our method with \btimerbaseline~and visualize five views from generated scenes. We observe significantly fewer artifacts and higher fidelity.
}
\label{fig:sota_static_comparison}
\end{figure}

For the task of image to 3DGS generation, we show qualitative comparisons with \btimerbaseline~in Fig.~\ref{fig:sota_static_comparison} and qualitative results in Tab.~\ref{tab:sota_comparison_static_lyra}. We observe higher quality and less artifacts for our method.

\subsection{Dynamic 3D Evaluation}
\label{sec:evaluation_4d}

\inlinesection{\btimerbaseline~comparison.}
For dynamic 3D evaluation, we compare our method with \btimerbaseline, as discussed in Sec.~\ref{sec:baseline_details}, on 100 out-of-distribution videos from our dynamic \methodname\ dataset. We crop the video to $512 \times 512$ for fair comparisons, as BTimer was mainly trained on that resolution. We show results in Tab.~\ref{tab:sota_comparison_dynamic} and observe that our method significantly outperforms \btimerbaseline. 

%\inlinesection{Dynamic data augmentation.}
%We conduct an ablation study by comparing our model with a variant without the dynamic data augmentation introduced in Sec.~\ref{sec:method_4d}. We evaluate on the full video resolution using extreme novel viewpoints to demonstrate the importance of the dynamic data augmentation. We show results in Tab.~\ref{tab:ablations_dynamic}

\begin{table}[t]
\centering
\small
\setlength{\tabcolsep}{4pt}

\begin{subtable}{0.48\linewidth}
\centering
\begin{tabular}{lccc}
\toprule
Method & PSNR~$\uparrow$ & SSIM~$\uparrow$ & LPIPS~$\downarrow$ \\
\midrule
\btimerbaseline & \textnormal{16.32} & \textnormal{0.580} & \textnormal{0.427} \\
Ours         & \textbf{24.92}     & \textbf{0.834}     & \textbf{0.183} \\
\bottomrule
\end{tabular}
\caption{\textbf{Comparison on static \methodname~dataset.}}
\label{tab:sota_comparison_static_lyra}
\end{subtable}
\hfill
\begin{subtable}{0.48\linewidth}
\centering
\begin{tabular}{lccc}
\toprule
Method & PSNR~$\uparrow$ & SSIM~$\uparrow$ & LPIPS~$\downarrow$ \\
\midrule
\btimerbaseline & \textnormal{20.29} & \textnormal{0.687} & \textnormal{0.315} \\
Ours         & \textbf{23.07}     & \textbf{0.779}     & \textbf{0.231} \\
\bottomrule
\end{tabular}
\caption{\textbf{Comparison on dynamic \methodname~dataset.}}
\label{tab:sota_comparison_dynamic}
\end{subtable}

\caption{\textbf{Quantitative results on static and dynamic \methodname~datasets.}}
\label{tab:sota_btimer}
% \vspace{-1em}
\end{table}

\subsection{Depth Visualization}
\label{sec:depth_visualization}

\begin{figure}[h]
\centering
\includegraphics[width=\textwidth]{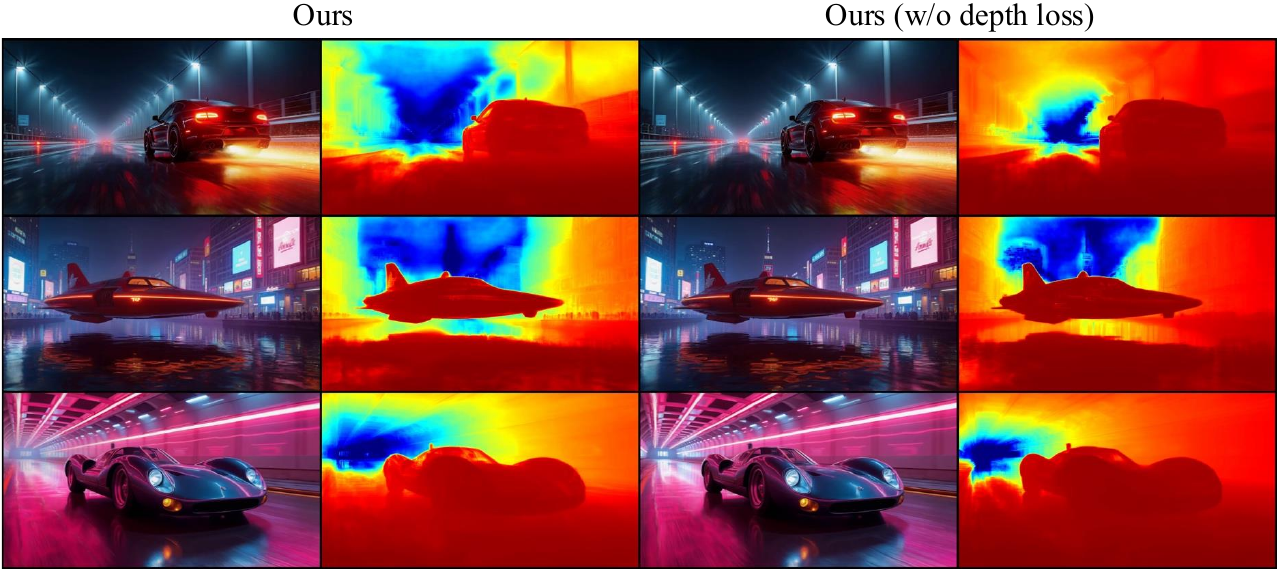}
\caption{\textbf{Depth loss ablation.} We visualize 3DGS renderings and corresponding depth maps. Using the depth loss as additional supervision prevents flat geometries without sacrificing visual quality.}
\label{fig:ablations_depth}
\vspace{-0.0cm}
\end{figure}

We visualize depths from our generated 3DGS in Fig.~\ref{fig:ablations_depth}. Using depth supervision prevents flat geometries without sacrificing visual quality.

\section{Robot Simulation}
\label{sec:method_robot}
\begin{figure}[t!]
\centering
\includegraphics[width=1.0\textwidth]{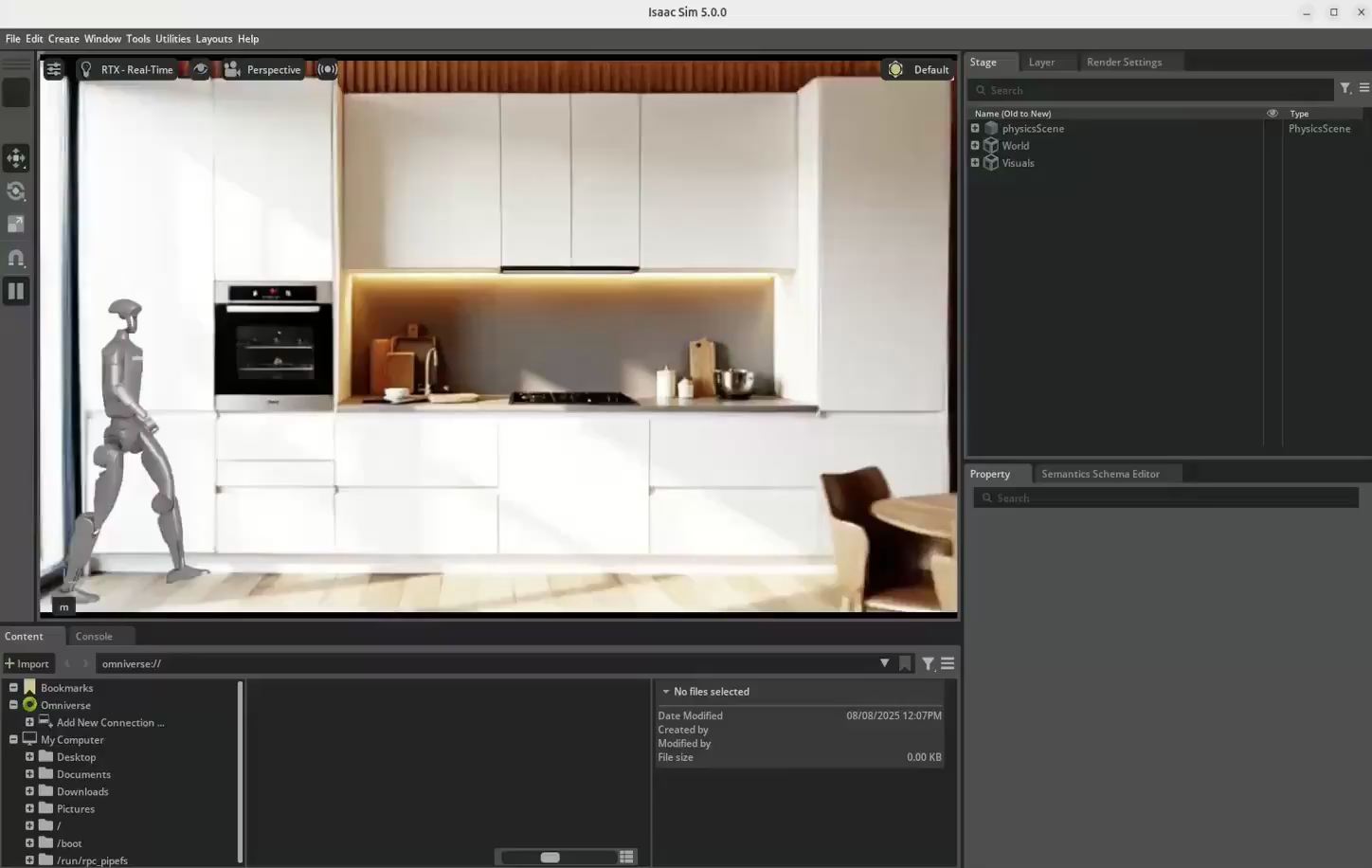}
\caption{\textbf{Robot Simulation.} We visualize a frame of a robot simulation within the Isaac Sim 5.0 simulation framework that takes a generated 3DGS scene from our method as input.
}
\label{fig:isaac}
\end{figure}

One of our primary motivations for building a diverse 3D generator is its ability to synthesize simulation environments for autonomous agents. To this end, we develop an end-to-end pipeline: we first generate 3D scenes from text with our model, export the 3D Gaussians as a .ply file, and convert the .ply files into a .usdz format. We use 3DGUT~\citep{wu20253dgut} and its export function to create .usdz files. The resulting .usdz file can then be imported into the NVIDIA Isaac robot development platform as a physically based virtual environment. AI-based robots can be trained and tested under diverse conditions within these environments. We present early demo results on our supplementary webpage and a screenshot of the interface in Fig.~\ref{fig:isaac}.

\begin{figure}[t]
\centering
\includegraphics[width=\textwidth]{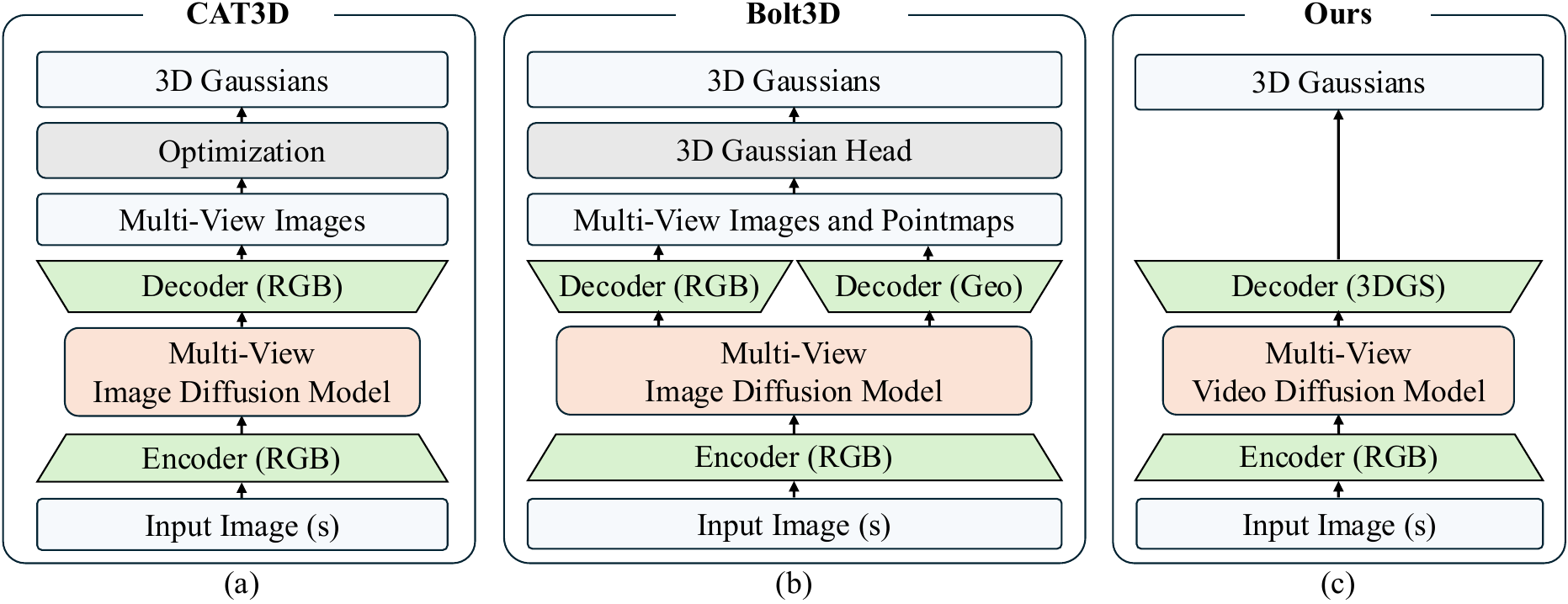}
\caption{\textbf{Approaches for 3D Generation.}
CAT3D~\citep{gao2024cat3d} proposed a multi-view image diffusion model that outputs images from novel viewpoints; subsequently, the images are reconstructed into 3D Gaussians using optimization. Bolt3D~\citep{szymanowicz2025bolt3d} fine-tunes CAT3D to output pointmaps using a geometry autoencoder; 
instead of relying on optimization, the 3D scene is predicted with a feed-forward 3D Gaussian head.
In contrast, our work builds upon a pre-trained camera-controlled video diffusion model and \textit{directly} decodes the multi-view video latents into 3D Gaussians.
}
\label{fig:intro_approaches}
\end{figure}

\section{Related Work}

In this section, we will discuss additional related work. We highlight the differences to previous works CAT3D~\citep{gao2024cat3d} and Bolt3D~\citep{szymanowicz2025bolt3d} in Fig.~\ref{fig:intro_approaches}.

\inlinesection{3D generation.}
Early approaches to 3D generation primarily focused on single object categories, extending GANs into the 3D domain by leveraging neural renderers as an inductive bias~\citep{devries2021unconstrained, chan2022efficient, or2022stylesdf, schwarz2022voxgraf, bahmani2023cc3d}. With the introduction of CLIP-based supervision~\citep{radford2021learning}, the field advanced toward more flexible and diverse asset creation, enabling both text-driven generation and editing~\citep{chen2018text2shape, jain2022zero, sanghi2022clip, jetchev2021clipmatrix, gao2023textdeformer, wang2022clip}. More recently, diffusion models have significantly improved quality by replacing CLIP guidance with Score Distillation Sampling (SDS)~\citep{poole2022dreamfusion, wang2023prolificdreamer, lin2022magic3d, chen2023fantasia3d, liang2023luciddreamer, wang2023score, li2024controllable, he2024gvgen, ye2024dreamreward, liu2023humangaussian, yu2023text, katzir2023noise, lee2023dreamflow, sun2023dreamcraft3d}.

To further enhance structural coherence, several methods enforce multi-view consistency by generating scenes from different perspectives~\citep{lin2023consistent123, liu2023zero, shi2023mvdream, feng2024fdgaussian, liu2024isotropic3d, kim2023neuralfield, voleti2024sv3d, hollein2024viewdiff, tang2024pixel, gao2024cat3d,wang2025act,kant2025pippo,yuan2025generative,ren2024xcube}, while others adopt iterative inpainting as a strategy for scene synthesis~\citep{hollein2023text2room, shriram2024realmdreamer}. Another line of work explores lifting 2D images into 3D using NeRF~\citep{NeRF}, 3D Gaussian Splatting~\citep{kerbl20233d}, or mesh-based representations coupled with diffusion models~\citep{chan2023generative, tang2023make, gu2023nerfdiff, liu2023syncdreamer, yoo2023dreamsparse, tewari2024diffusion, qian2023magic123, long2023wonder3d, wan2023cad, szymanowicz2023viewset,lu2024infinicube}.

\inlinesection{Feed-forward 3D models.}
Complementary efforts investigate fast, feed-forward techniques that directly predict 3D content from images or text inputs~\citep{hong2023lrm, li2023instant3d, xu2023dmv3d, xu2024grm, zhang2024compress3d, han2024vfusion3d, jiang2024brightdreamer, xie2024latte3d, tang2024lgm, tochilkin2024triposr, qian2024atom, szymanowicz2023splatter, szymanowicz2024flash3d, liang2024wonderland, szymanowicz2025bolt3d,schwarz2025generative,yang2025matrix,zhang2025spatialcrafter}. However, in contrast to our method, these approaches remain limited to producing static 3D scenes. Other methods focus on specific scenes such as faces~\citep{kirschstein2025avat3r}. Concurrent work~\citep{liang2024btimer,xu20254dgt} tackles real scenes but can not handle diverse generated scenes or large viewpoint changes.

\inlinesection{4D generation.}
Recent years have seen rapid progress in 4D generation, i.e., dynamic 3D scene synthesis. Many approaches leverage input text prompts or images as guidance. Following the introduction of large-scale generative models for this task~\citep{singer2023text}, subsequent works have made notable advances in both visual fidelity and motion quality~\citep{ren2023dreamgaussian4d, ling2023align, bahmani20234d, zheng2023unified, gao2024gaussianflow, yang2024beyond, jiang2023consistent4d, miao2024pla4d, li2024vivid, zhang4900422motion4d, yuan20244dynamic, jiang2024animate3d}.

While text conditioning is widely used, alternative methods aim to lift 2D images or videos into dynamic 3D scenes~\citep{ren2023dreamgaussian4d, zhao2023animate124, yin20234dgen, pan2024fast, zheng2023unified, ling2023align, gao2024gaussianflow, zeng2024stag4d, chu2024dreamscene4d, wu2024sc4d, yang2024diffusion, wang2024vidu4d, feng2024elastogen, sun2024eg4d, zhang20244diffusion, ren2024l4gm, lee2024vividdream, li20244k4dgen, van2024generative, uzolas2024motiondreamer, chai2024star, liang2024diffusion4d, li2024dreammesh4d, SV4D,li2025fb,wang2025video4dgen,zhu2025ar4d,zhou2025coco4d,hu2025ex,park2025zero4d}. Other efforts explore incorporating physics priors into generation pipelines~\citep{lin2024phy124, huang2024dreamphysics, zhang2024physdreamer,kiray2025promptvfx}, or enhancing motion controllability with template-driven approaches~\citep{zhang2024magicpose4d, sun2025ponymation, chen2024ct4d}. A parallel direction emphasizes compositional and interactive 4D generation~\citep{bahmani2024tc4d, xu2024comp4d, cao2024avatargo, yu20244real, zeng2024trans4d, zhu2024compositional}.

Another strand of research extends 3D GAN frameworks into the 4D setting by training directly on 2D video data~\citep{bahmani20223d, xu2022pv3d}. However, such methods are typically limited by small-scale, single-category datasets, which constrain generalization. Moreover, most existing approaches remain object-centric, often neglecting background elements. As a result, their overall visual fidelity lags behind the high photorealism achieved by recent video generation models, which our method leverages.

\inlinesection{Camera-conditioned video models.}
Recent progress has focused on incorporating camera control into video diffusion models. The pioneering work MotionCtrl~\citep{MotionCtrl} introduced camera conditioning by augmenting pre-trained video models~\citep{VideoCrafter1, blattmann2023stable} with extrinsic matrices. Subsequent efforts~\citep{CameraCtrl, xu2024camco, kuang2024collaborative,ju2025fulldit,he2025cameractrl,li2025realcam,wang2025akira,wang2024cpa,lei2025animateanything,liu2025idcnet} enhanced conditioning by representing cameras using Plücker coordinates, while others use follow MotionCtrl to simply use poses flattened~\citep{wang2025cinemaster}. Another line of research~\citep{MotionMaster, xiao2024video, ling2024motionclone, hou2024training} enabled camera motion control without introducing additional trainable parameters. Despite these advances, all of the above rely on U-Net-based architectures. Other approaches use low-rank approaches for camera control~\citep{zhang2025lion}.

Several recent works continue to extend U-Net-based approaches for camera control~\citep{zhao2024genxd, xu2024cavia, zheng2024cami2v, zhang2024recapture, yu2024viewcrafter, lei2024animateanything}, while others begin exploring diffusion transformers. Cheong et al.\citep{cheong2024boosting} propose one such transformer-based framework, though its scene and visual quality remain limited. DimensionX\citep{sun2024dimensionx} introduces joint space-time control in diffusion transformers, but relies on pre-defined, non-continuous camera trajectories. Other works~\citep{CamPoseVDiT,mai2025can,jiang2025geo4d} instead explore pose and geometry estimation with a video DiT. CAT4D~\citep{wu2024cat4d} proposes a multi-view video diffusion model derived from fine-tuning a multi-view backbone. Large-scale synthetic data generation further boosts performance, as demonstrated by SynCamMaster~\citep{bai2024syncammaster} and ReCamMaster~\citep{bai2025recammaster}, which achieve strong results on camera-controlled video generation using transformer-based architectures. Finally, 4DiM~\citep{4DiM} trains a space-time diffusion model from scratch for novel view synthesis from a single input image.
Another line of work edits camera movements of videos with underlying motion using camera retargeting~\citep{seo2025vid,ma2025follow}.
One popular direction has been to use a spatio-temporal cache, e.g., using point clouds or optical flow, to condition the camera-controlled video generation~\citep{wu2025video,ren2025gen3c,wang2025longdwm,cao2025uni3c,yu2025trajectorycrafter,jin2025flovd,xing2025motioncanvas,yan2025streetcrafter,li2025martian,gu2025diffusion}.

\end{document}